\documentclass[letterpaper, 10 pt, conference]{ieeeconf}  

\usepackage{amsmath}
\usepackage{adjustbox}
\usepackage{graphicx,float}
\usepackage{epstopdf}
\usepackage{caption}
\usepackage{pdflscape}
\usepackage{amssymb}
\usepackage{amsfonts}
\usepackage{pgfplots}
\usepackage{bbm}
\usepackage{color}
\usepackage{multirow}
\usepackage{tikz}
\usepackage{dblfloatfix}
\usepackage{float}
\usepackage{hyperref}
\usepackage{adjustbox}
\usepackage{balance}
\usepackage{algorithm}
\usepackage{algpseudocode}
\usepackage{relsize}
\usepackage{tabularx} 
\usepackage[compress]{cite}

\usepackage{verbatim}  

\makeatletter
\newcommand*\titleheader[1]{\gdef\@titleheader{#1}}
\AtBeginDocument{%
	\let\st@red@title\@title
	\def\@title{%
		\bgroup\normalfont\large\centering\@titleheader\par\egroup
		\vskip1.5em\st@red@title}
}
\makeatother

\IEEEoverridecommandlockouts                              

\titleheader{This work has been submitted to the IEEE for possible publication. Copyright may be transferred without notice, after which this version may no longer be accessible.}

\title{\LARGE \bf
Impact of Temporal Delay on Radar-Inertial Odometry
}

\author{Vlaho-Josip Štironja$^{1}$, Luka Petrović$^{1}$, Juraj Peršić$^{2}$, Ivan Marković$^{1}$,  and Ivan Petrović$^{1}$
\thanks{$^{1}$University of Zagreb Faculty of Electrical Engineering and Computing, Laboratory for Autonomous Systems and Mobile Robotics (LAMOR), Unska 3, HR-10000, Zagreb, Croatia
        {\tt\small \{name.surname\}@fer.unizg.hr}}%
\thanks{$^{2}$Calirad d.o.o.,
Augusta Harambašića 4, HR-10000, Zagreb, Croatia
        {\tt\small  juraj.persic@calirad.net}}%
}

\begin{document}

\maketitle
\thispagestyle{empty}
\pagestyle{empty}

\begin{abstract}
Accurate ego-motion estimation is a critical component of any autonomous system.
Conventional ego-motion sensors, such as cameras and LiDARs, may be compromised in adverse environmental conditions, such as fog, heavy rain, or dust. 
Automotive radars, known for their robustness to such conditions, present themselves as complementary sensors or a promising alternative within the ego-motion estimation frameworks.
In this paper we propose a novel Radar-Inertial Odometry (RIO) system that integrates an automotive radar and an inertial measurement unit. 
The key contribution is the integration of online temporal delay calibration within the factor graph optimization framework that compensates for potential time offsets between radar and IMU measurements.
To validate the proposed approach we have conducted thorough experimental analysis on real-world radar and IMU data.
The results show that, even without scan matching or target tracking, integration of online temporal calibration significantly reduces localization error compared to systems that disregard time synchronization, thus highlighting the important role of, often neglected, accurate temporal alignment in radar-based sensor fusion systems for autonomous navigation.
Project website: \href{https://rio-online-t.github.io/}{https://rio-online-t.github.io/}.
\end{abstract}

\section{Introduction}
Accurate and robust localization is critical for autonomous navigation, especially in challenging environments such as urban areas with tall buildings or tunnels where GNSS signals are not available, as well as in adverse weather conditions that affect the performance of cameras and LiDARs. 
While cameras and LiDARs are commonly used for ego-motion estimation, their performance degrades in poor visibility, heavy rain or dense fog, whereas automotive radars, with their longer operating wavelengths, provide a robust alternative that can withstand the aforementioned environmental conditions \cite{review}. 
In addition, radars have the advantage of providing highly accurate and robust ego-velocity estimation using Doppler shift measurements, as shown in \cite{rave, kellner_R}, and can be integrated into tightly coupled sensor fusion ego-motion estimators, where it is assumed that each radar-seen object in the environment is static, so that it is possible to establish a relationship between the Doppler velocity of the points and the ego-velocity of the sensor.

Despite its advantages, radar-based ego-motion estimation faces several challenges, including lower spatial resolution, multipath reflections, and cluttered measurements.
On the other hand, while IMUs provide accurate angular velocity measurements, the presence of sensor biases leads to drift in odometry over time.
This complementary nature makes radar-IMU fusion a promising approach for ego-motion estimation. 
In contrast to vision- or LiDAR-based methods, this fusion remains unaffected by environmental conditions, making it a robust solution for autonomous navigation under challenging perceptual conditions.
The increasing interest in radar-IMU fusion has recently accelerated research in this area, leading to state-of-the-art advancements in radar-inertial ego-motion estimation \cite{mfi-rik, less_is_more, dero, multi_radar_cauchy, michalcyzk_2024, coral, riv, unc_2025}, among many others.

\begin{figure}[t]
    \centering
    \includegraphics[width=0.47\textwidth]{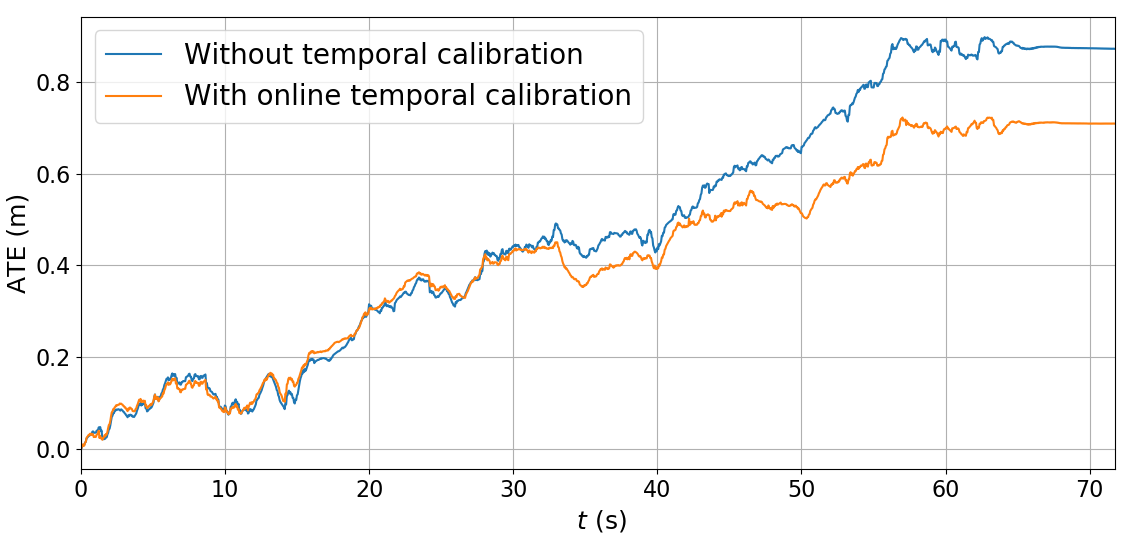}
    \caption{Absolute trajectory error (m) with origin alignment on the Mocap difficult sequence from the IRS dataset~\cite{irs}, comparing the loosely coupled factor graph-based radar-inertial odometry without our proposed modification and the same approach with our proposed modification of the factor graph structure. The results highlight the positive effects of integrating online temporal calibration and emphasize the overall importance of accurate temporal calibration for improved odometry performance.}
    \label{fig::first_page}
    \vspace{-6mm}
\end{figure}

Doer et al.~\cite{doer_ekf} integrated ego-velocity estimates from 3D radar together with IMU and barometer data within an Error State Extended Kalman Filter (ES-EKF) framework to mitigate drift.
In their follow-up work~\cite{doer_online}, the authors introduced radar online extrinsic calibration, improving system accuracy without the need for tedious manual calibration.
Michalczyk et al.~\cite{jan_first} proposed a 3D tightly coupled EKF-based radar-inertial odometry method that includes range measurements to mitigate drift, reducing reliance on point angle information, which can be noisy in low-cost Frequency Modulated Continuous Wave (FCMW) radars.
In their extended work~\cite{jan_multi}, the authors introduced persistent landmark tracking across robot poses, significantly reducing the final odometry drift.
The authors also showed that addition of the extrinsic calibration parameters in the state (online calibration) improves the accuracy and consistency of the estimation~\cite{jan_online}.
DRIO~\cite{DRIO} introduced a novel radar-inertial odometry method that improves localization accuracy by using ground points as ever-present static references that are unaffected by dynamic objects, making it well-suited for highly dynamic environments.

The first 3D radar-inertial odometry and mapping system was introduced in~\cite{iriom}, where the authors used an iterative EKF and adopted a graduated non-convexity method for outlier rejection instead of the commonly used RANSAC.
The authors in~\cite{do_we_need_sm} analyzed the error-prone nature of radar's scan-to-scan and scan-to-map registration algorithms, comparing them with Doppler/IMU integration and EKF sensor fusion.
They demonstrated that sensor fusion produced results comparable to or better than radar's 3D point matching.
Similarly, in \cite{movro2} it was shown that the  radar-intertial odometry  scan matching did not significantly improve accuracy when using a low-cost automotive radar~\cite{movro}. 
However, scan matching expanded the system's capabilities, in the sense that if monocular odometry failed, ego-motion estimation could still be reliably corrected through the scan matching.
Furthermore, in~\cite{unc_2025} authors highlighted the potential of integrating radar point uncertainty to enhance radar Simultaneous Localization and Mapping (SLAM) systems in adverse weather conditions.
DeRO odometry~\cite{dero} used Doppler velocity in conjunction with gyroscope data to estimate poses, mitigating high drift caused by accelerometer biases and double integration.

The authors in \cite{kelly2021question} point out fundamental structural problems when using recursive filters such as the EKF for temporal calibration, which can lead to distortions and inconsistencies despite careful tuning. 
To overcome these problems, we use a factor graph-based approach where the offset is estimated online while maintaining the flexibility to adjust the size of the optimization window to cover every possible delay interval, as proposed in  \cite{kelly2021question}, or to compute it offline using the entire graph. 
As shown in \cite{michalcyzk_2024} for the case of radar-inertial odometry, the authors compared an approach to optimize the factor graph with a multi-state EKF-based RIO. 
While the EKF-based method requires fewer resources and performs similarly well to the factor graph approach when the linearization point is close to the true state, the factor graph method shows better convergence during transient phases, especially when the linearization point is far from the true state due to poor initialization or when the uncertainty of the initial state does not reflect the true uncertainty. 
Considering these advantages and the flexibility that factor graphs offer for possible future extensions, we have opted for this formulation.

Inspired by the improved odometry accuracy and the convergence of extrinsic parameters achieved by integrating extrinsic calibration parameters into the state and performing online calibration in \cite{jan_online, doer_online} compared to previous approaches \cite{jan_first, doer_ekf}, we extend this idea by taking into account the time delay between the radar and the IMU. 
Existing approaches often overlook the time lag and assume perfect synchronization, which limits both accuracy and robustness in real-world scenarios. To address this limitation, in this work we propose a novel radar-inertial odometry based on factor graph optimization that integrates a temporal offset factor into the state to account for time delays between radar and IMU messages in real time. 
We deliberately chose not to include an extrinsic online calibration in order to investigate the direct effects of the temporal offset calibration.

Experimental results on a real-world dataset demonstrate that accounting for temporal offset increases the accuracy of ego-motion and provides competitive results even without scan matching or target tracking.
Introducing this temporal offset correction into other methods can also further reduce localization and mapping errors, especially in dynamic environments (see Fig. 1). 
Furthermore, we tested the convergence of our temporal delay estimator by simulating different radar timestamps to evaluate whether our estimator adapts to timestamp variations. 
Finally, our approach achieves competitive performance compared to state-of-the-art solutions, emphasizing the importance of considering time delays in radar-inertial sensor fusion, especially in agile motion scenarios.

Summarized are our main contributions:
\begin{itemize}
    \item A factor graph-based radar-inertial odometry that integrates online temporal calibration, denoted as RIO-T.
    \item A demonstration of the consistency of the temporal offset estimation under different timestamp variations.
    \item An analysis of the effect of introducing a temporal delay within the radar-IMU sensor fusion estimators in factor graph optimization on localization accuracy, as well as the overall effect of temporal offset on ego-motion estimation.
\end{itemize}

\section{Proposed Radar-Inertial Odometry Architecture}
\label{sec:structure}

\begin{figure}[t]
    \centering
    \includegraphics[width=0.5\textwidth]{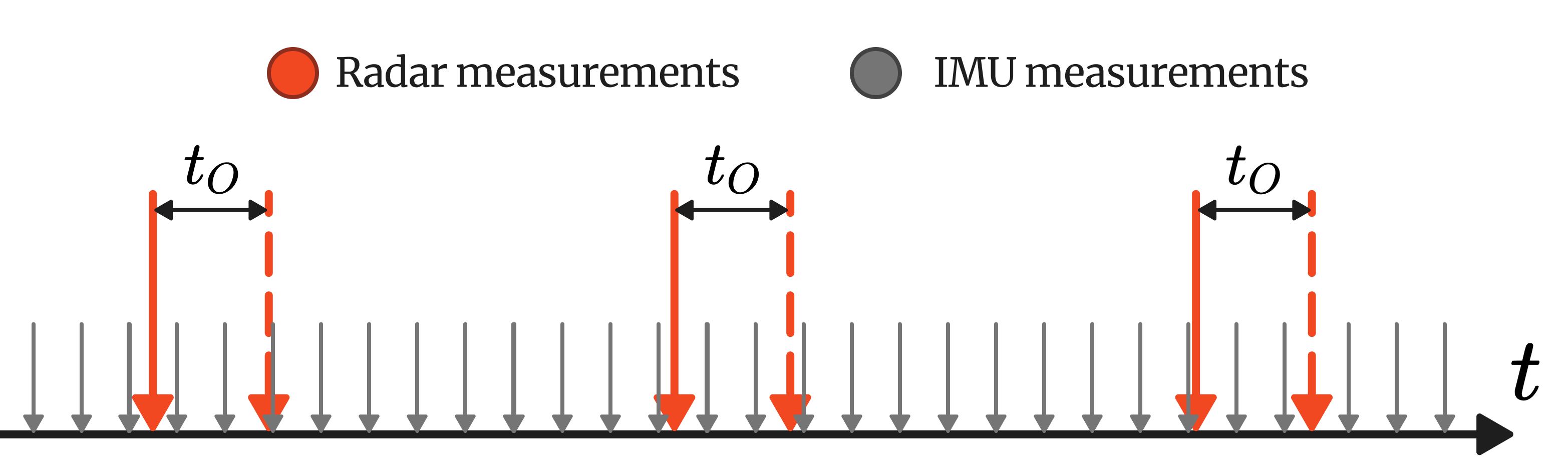}
    \caption{An illustration of a radar-inertial odometry system with unsynchronized measurements. We estimate the temporal offset between radar measurements and IMU measurements and optimize it within the factor graph framework. Radar ego-velocity measurements (red) indicate the time at which radar data is represented, while the red dashed line marks radar sequence timestamps. IMU measurements are shown in grey. }
    \label{fig::time}
    \vspace{-5mm}
\end{figure}

The proposed method, dubbed RIO-T (Radar-Inertial Odometry with Temporal Offset Calibration), estimates odometry using a loosely coupled radar-inertial sensor fusion approach implemented within a factor graph optimization framework.
The method imposes no assumptions on the environment, such as the presence of planar surfaces, i.e., ground plane estimation.
The core idea is to adjust the radar ego-velocity factor by compensating for the velocity difference using the most recent IMU acceleration, with having the bias removed and measurement corrected for gravity, and scaled by the time-offset state.
The assumption of constant acceleration around the last acceleration measurement provides a robust and accurate real-time estimate of ego motion while estimating the time lag between radar and IMU measurements, as shown in Fig.~\ref{fig::time}. 
The RIO-T method is described in detail in the following sections.

\vspace{-1mm}
\subsection{Notation, state definition, and assumptions}
\vspace{-0.5mm}

The following notation is used in this paper.
Scalars are written in non-bold lowercase letters, e.g., $m$.
Constants are written in non-bold capital letters, e.g., $\mathrm{M}$, and matrices in bold capital letters, e.g., $\mathbf{M}$.
Vectors are written in bold lower case letters, e.g. $\mathbf{m}$.
The spelling of rotation and vector is taken from~\cite{notation}.
The expression ${}_{\mathrm{{{{\scalebox{0.56}{A}}}}}}\mathbf{{\mkern-1.7mu t}_{{{\scalebox{0.56}{BC}}}}}$ stands for the translation vector from frame $\mathcal{F}_{{\scalebox{0.56}{B}}}$ to $\mathcal{F}_{{\scalebox{0.56}{C}}}$, expressed in frame $\mathcal{F}_{{\scalebox{0.56}{A}}}$.
The rotation matrix $\mathbf{{R}_{{{\scalebox{0.56}{AB}}}}}$ is the rotation of the frame $\mathcal{F}_{{\scalebox{0.56}{B}}}$ with respect to $\mathcal{F}_{{\scalebox{0.56}{A}}}$, i.e., $\mathbf{{R}_{{{\scalebox{0.56}{AB}}}}}$ transforms a vector ${}_{\mathrm{{{{\scalebox{0.56}{B}}}}}}\mathbf{{\mkern-1.7mu v}}$ from $\mathcal{F}_{{\scalebox{0.56}{B}}}$ to $\mathcal{F}_{{\scalebox{0.56}{A}}}$ (${}_{\mathrm{{{{\scalebox{0.56}{A}}}}}}\mathbf{{\mkern-1.7mu v}}$).

The system state at step $k$ is defined as:
\begin{equation}
    \mathbf{x}_k = \begin{bmatrix}  
        \mathbf{R}_{\mathrm{\scalebox{0.56}{IW}},k} &  
        \mathrm{{}_{\scalebox{0.56}{W}}}\mathbf{p}_{\mathrm{\scalebox{0.56}{IW}},k} &  
        \mathrm{{}_{\scalebox{0.56}{W}}}\mathbf{v}_k &  
        \mathbf{b}_{\mathrm{\scalebox{0.56}{g}},k} &  
        \mathbf{b}_{\mathrm{\scalebox{0.56}{a}},k} &  
        t_{\mathrm{\scalebox{0.56}{O}},k}  
    \end{bmatrix},
    \label{state}
\end{equation}  
where $\mathbf{R}_{\mathrm{\scalebox{0.56}{IW}},k}$, $\mathrm{{}_{\scalebox{0.56}{W}}}\mathbf{p}_{\mathrm{\scalebox{0.56}{IW}},k}$, $\mathrm{{}_{\scalebox{0.56}{W}}}\mathbf{v}_k$, $\mathbf{b}_{\mathrm{\scalebox{0.56}{g}},k}$, $\mathbf{b}_{\mathrm{\scalebox{0.56}{a}},k}$, and $t_{\mathrm{\scalebox{0.56}{O}},k}$ represent the orientation, position, velocity, gyroscope bias, accelerometer bias, and temporal offset at step $k$, respectively.  
With each new measurement, an additional system step is added to the state. 
The extrinsic parameters, i.e., the transformation between the IMU and radar frames, consist of the translation component $\mathrm{{}_{\scalebox{0.56}{I}}}\mathbf{t}_{\mathrm{\scalebox{0.56}{RI}}}$ and the orientation component $\mathbf{R}_{\mathrm{\scalebox{0.56}{RI}}}$, both assumed known a priori.  
The frames $\mathcal{F}_{\scalebox{0.56}{I}}$ and $\mathcal{F}_{\scalebox{0.56}{W}}$ correspond to the IMU and world frames, respectively, with the world frame being initialized at the start of the system. 
Orientations are represented as rotation matrices $\mathbf{R} \in \mathrm{SO}(3)$~\cite{barfoot}.  
Since the IRS dataset sequences are static at the start, we estimate the gyroscope bias and initial pitch and roll angles, while setting the acceleration bias to zero. 
The mobile platform is assumed to be stationary, with position, velocity, and temporal offset $t_{\mathrm{\scalebox{0.56}{O}},k} $ initialized to zero, while the initial state uncertainty remains a tunable parameter. 
The extrinsic calibration between IMU and radar ($\mathbf{t}_{\mathrm{\scalebox{0.56}{RI}}}$,  $\mathbf{R}_{\mathrm{\scalebox{0.56}{RI}}}$) is assumed to be known.

\vspace{-1mm}
\subsection{Factor graph model and factors}
\vspace{-0.5mm}
Factor graph optimization (FGO) for multi-sensor fusion is formulated as a maximum a posteriori (MAP) estimation problem~\cite{barfoot}, where the system state is optimized using probabilistic constraints from measurements and prior knowledge~\cite{gtsam_intro}. 
Assuming conditional independence of measurements, the MAP problem is expressed as:
\begin{equation}
    \hat{\mathbf{x}}=\underset{\mathbf{x}}{\mathrm{argmax}} \prod_{k,i}p_i(\mathbf{z}_{k,i} | \mathbf{x}_k) \prod_{k}p(\mathbf{x}_k| \mathbf{x}_{k-1},\mathbf{u}_k),
\label{fgo_1}
\end{equation}
where $\mathbf{x}_k$ represents the system state at time $k$, while $\mathbf{z}_{k,i}$ and $\mathbf{u}_k$ denote the measurements from $i$-th sensor and control inputs, respectively. 
By assuming Gaussian noise with zero mean and covariance $\Sigma_k$, and that all nodes are independent, the factor graph optimization problem can be formulated as a nonlinear least squares minimization:
\begin{equation}
     \hat{\mathbf{x}} = \underset{\mathbf{x}}{\mathrm{argmin}} \sum_{k,i}\| \mathbf{z}_{k,i} - \mathbf{h}_{k,i}(\mathbf{x}_k) \|^2_{\Sigma_{k,i}},
     \label{fgo_4}
\end{equation}
where $\mathbf{h}_{k,i}(\cdot)$ represents the measurement function, and $\Sigma_{k,i}$ the covariance matrix associated with the $i$-th measurement at time $k$.

\begin{figure}[t]
    \centering
    \includegraphics[width=0.5\textwidth]{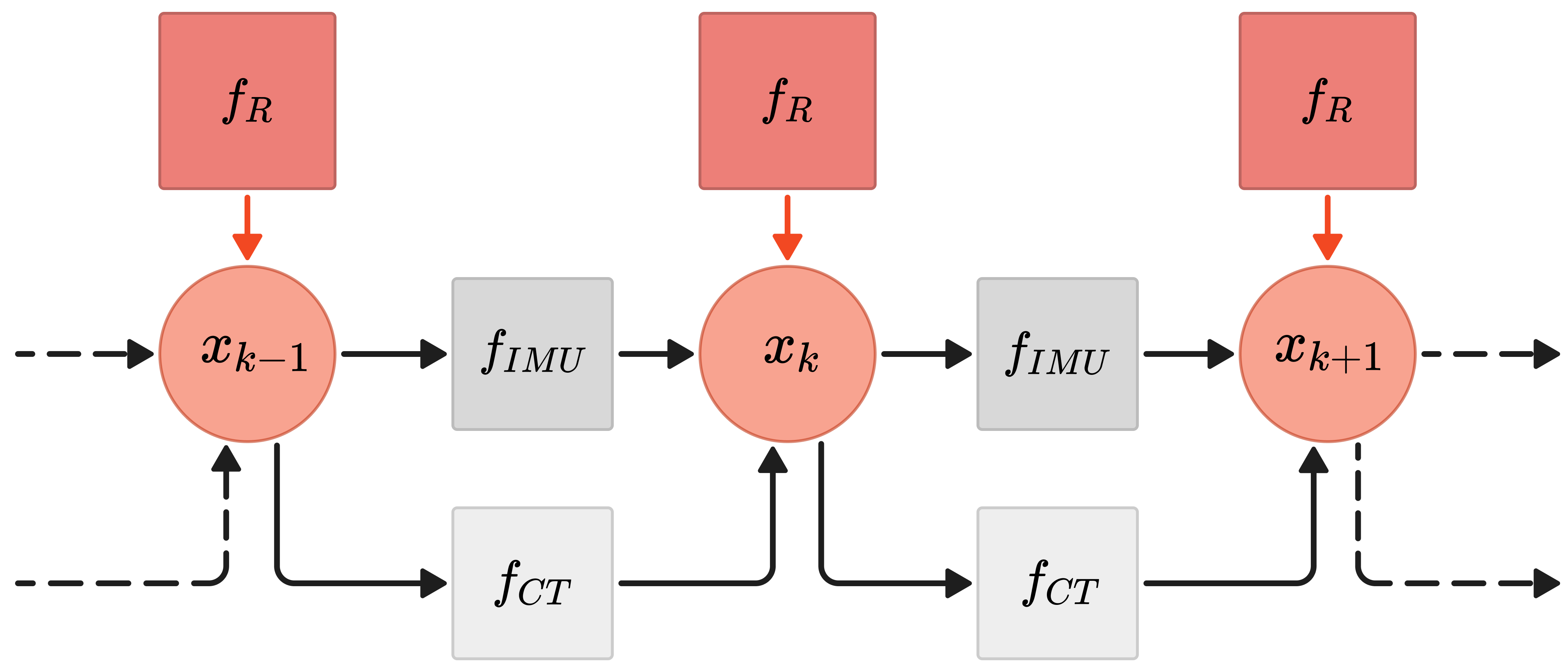}
    \caption{ Proposed factor graph structure.
The factor graph consists of the IMU factor ($f_{IMU}$), the radar ego-velocity factor ($f_{R}$), and constant time offset factor ($f_{CT}$).}
    \label{fig::fgo_structure}
    \vspace{-5mm}
\end{figure}

The architecture of the proposed RIO-T framework is depicted in Fig.~\ref{fig::fgo_structure}. 
The factor graph comprises three key components: the IMU preintegration factor ($f_{IMU}$), the radar ego-velocity factor ($f_{R}$), and the constant time offset factor ($f_{CT}$). 
The IMU and the time offset factors impose constraints between consecutive states, ensuring temporal consistency, while the radar ego-velocity factor provides velocity measurements to refine the state trajectory. 
Additionally, the radar factor compensates for temporal misalignment by dynamically adjusting the estimated time offset.

\subsubsection{The ego-velocity factor of the radar $f_R$}
For ego-velocity estimation, we opted to use the open-source package RAVE~\cite{rave}, using RANSAC for outlier rejection and least squares (LS) for state estimation. 
A sliding average filter is applied to discard infeasible estimates.
The measurement function $h^R$ for the ego-velocity factor of the radar $f_R$ is defined as:
\begin{equation}
    h^R (\mathbf{x}_{k}) = \mathbf{R}_{\mathrm{\scalebox{0.56}{RI}}} \left( \mathbf{R}_{\mathrm{\scalebox{0.56}{IW}}, k} \, {}_{{\scalebox{0.56}{W}}}\mathbf{v}_k + {}_{{\scalebox{0.56}{I}}}{\boldsymbol{\omega}}_k \times {}_{{\scalebox{0.56}{I}}} \mathbf{t}_{{{\scalebox{0.56}{RI}}}} 
    - t_{\mathrm{\scalebox{0.56}{O}},k} \cdot {}_{{\scalebox{0.56}{I}}}\mathbf{a}_{\mathrm{c},k}  \right),
    \label{hr_imu}
\end{equation}
where $\boldsymbol{\omega}_k$ and $\mathbf{a}_k$ denote the most recent IMU angular velocity and acceleration measurements prior to the radar data update.
To compute the corrected acceleration $\mathbf{a}_{\mathrm{c},k}$, we subtract the transformed gravity from the world frame, along with the acceleration state bias, from the current IMU acceleration measurement and express the result in the current frame as follows:

\begin{equation}
    {}_{{\scalebox{0.56}{I}}}\mathbf{a}_{\mathrm{c},k} =  {}_{{\scalebox{0.56}{I}}}\mathbf{a}_k - \mathbf{R}_{\mathrm{\scalebox{0.56}{IW}},k}\mathbf{g} -  \mathbf{b}_{\mathrm{\scalebox{0.56}{a}},k}.
    \label{acc_correction}
\end{equation}

Since the temporal offset $t_{\mathrm{\scalebox{0.56}{O}},k}$ is estimated online, and the assumption of constant acceleration near the radar measurement is an approximation, larger offsets may degrade the accuracy of temporal offset estimation. 
Nevertheless, the estimated offset can serve as an initial guess for more refined temporal calibration methods, enhancing their convergence.
The measurement model of the radar can be described as follows:
\begin{equation}
    \mathbf{z}_k = h^{R}(\mathbf{x}_{k}) + \mathbf{v}^{R}_{k},
\end{equation}
where $\mathbf{v}^{R}_{k}$ stands for Gaussian noise with zero mean and covariance $\sum_k^{R}$.
Moreover, we can define the error factor function as follows:
\begin{equation}
        \| \mathbf{e}^{R}_k \|^2_{\sum_k^{R}} = \|\mathbf{z}_{k} -h^{R}(\mathbf{x}_{k})  \|^2_{\sum_k^{R}}.
\end{equation}

\subsubsection{The constant time offset factor $f_{CT}$}

To constrain the consecutive nodes in the factor graph, we introduce the constant time offset factor $f_{CT}$.  
The measurement function $h^{CT}$ of this factor is expressed as follows:  
\begin{equation}
    h^{CT}(\mathbf{x}_{k-1}) = t_{\mathrm{\scalebox{0.56}{O}},k-1} .
    \label{hct}
\end{equation}
Therefore, the state at step $k$ can be described as follows:  
\begin{equation}
    \mathbf{x}_k = h^{CT}(\mathbf{x}_{k-1}) + \mathbf{v}^{CT}_{k-1},
\end{equation}  
where $\mathbf{v}^{CT}_{k-1}$ is zero-mean Gaussian noise with covariance $\sum_k^{CT}$.
We now define the error function ($\mathbf{e}^{CT}_k$) as:  
\begin{equation}
    \| \mathbf{e}^{CT}_k \|^2_{\sum_k^{CT}} = \|\mathbf{x}_{k} -h^{CT}(\mathbf{x}_{k-1})  \|^2_{\sum_k^{CT}}.
\end{equation}

\subsubsection{IMU factor $f_{IMU}$}
Including high-rate IMU measurements as individual factors would significantly increase the number of optimization variables, making real-time computation infeasible.
To address this, Forster et al.~\cite{imu_factor} introduced a preintegration method that combines multiple IMU measurements into a single factor. 
We have adopted this approach in our implementation, following~\cite{mfi-rik}.
To incorporate IMU measurements into the factor graph, we define the IMU error function $\mathbf{e}^{IMU}_k$ as:
\begin{equation}
    \| \mathbf{e}^{IMU}_k \|^2_{\sum_k^{IMU}} = 
    \| [\mathbf{e}_{\mathrm{R},k}, \mathbf{e}_{\mathrm{p},k}, \mathbf{e}_{\mathrm{v},k}, \mathbf{e}_{\mathrm{b_g},k},  \mathbf{e}_{\mathrm{b_a},k}]^\top \|^2_{\sum_k^{IMU}},
    \label{e_imu}
\end{equation}
where $\mathbf{e}_{\mathrm{R},k}$, $\mathbf{e}_{\mathrm{p},k}$, and $\mathbf{e}_{\mathrm{v},k}$ represent the errors in rotation, position, and velocity, respectively, based on the IMU kinematic integration, while $\mathbf{e}_{\mathrm{b_g},k}$ and $\mathbf{e}_{\mathrm{b_a},k}$ ensure consistency in gyroscope and accelerometer bias estimation.
The rotation error $\mathbf{e}_{\mathrm{R},k}$ is computed as the logarithm map of the product between the transpose of the preintegrated rotation and the estimated relative rotation between time steps $k-1$ and $k$ in the Lie algebra $\mathfrak{so}(3)$.

We can now formulate the  factor graph optimization problem, where the total error function is calculated as the sum of the error functions of all factors across the states in the graph.
Due to sensor noise and possible outliers in radar velocity measurements and time offset estimates, direct optimization can be sensitive to erroneous estimates. 
To enhance robustness, we use a Huber loss kernel, following the approach of Huang et al.~\cite{multi_radar_cauchy}. 
The optimal state $\hat{\mathbf{x}}$ is then obtained by solving the following factor graph optimization problem:
\begin{equation}
 \begin{array}{l}
    \hat{\mathbf{x}} = \underset{\mathbf{x}}{\mathrm{argmin}}\mathlarger{\sum_k} \Bigl( \| \mathbf{e}^{IMU}_k \|^2_{\sum_k^{IMU}} + \| \mathbf{e}^{CT}_k \|^2_{\sum_k^{CT}} \\ 
    + \rho \bigl( \| \mathbf{e}^{R}_k \|^2_{\sum_k^{R}} \bigl)  \Bigl).
    \label{fgo_full}
      \end{array}
\end{equation}
Similar to \cite{mfi-rik}, we solve the optimization problem using the GTSAM library~\cite{gtsam_perception} with a two-thread architecture based on the iSAM2 solver. 
The optimization thread solves the MAP problem with a 1-second data window, while the navigation thread outputs the last state prediction at the IMU rate.
To ensure a fair evaluation of the online performance of our proposed algorithm, we compare the odometry results obtained by the navigation thread.

\section{Experimental results}
\label{sec:results}
In this section, we present the results of the experimental validation of our method on several sequences from the IRS dataset~\cite{irs}. 
For comparison with state-of-the-art radar-inertial odometry approaches, we also evaluated state-of-the-art EKF-based and FGO-based radar-inertial ego-motion estimators and analyzed their accuracy with and without our estimated temporal offset to investigate the effects of temporal misalignment. 
Both the Relative Pose Error (RPE) and the Absolute Trajectory Error (ATE) with origin alignment were calculated with the open-source package~\cite{evo}. 
The RPE provides information on the short-term consistency of the estimated motion, which is crucial for applications requiring local accuracy, while the ATE quantifies the overall drift over time.
The IRS dataset includes measurements from a high-performance IMU (Analog Devices ADIS16448), an FMCW radar (Texas Instruments IWR6843AOP) and a monocular grayscale camera (IDS UI-3241), all synchronized via a hardware trigger and synchronization signals from a microcontroller board. 
Given this synchronization setup, we do not expect any significant deviations in the temporal offset of the radar, and for all scenarios using our proposed method, we initialize the temporal offset to 0.0 ms. 
However, as we will show later, this assumption may not even hold for a hardware-synchronized radar.
To thoroughly evaluate the robustness and accuracy of our approach, we performed three analyzes: (1) investigating the convergence behavior of the estimated temporal offset, (2) comparing the odometry accuracy between our proposed method and other state-of-the-art radar- inertial odometry approaches, and (3) investigating the impact of the temporal offset on the downstream radar- inertial odometry by testing algorithms with radar timestamps artificially shifted by the delay estimated by our method.

\begin{figure}[t]
    \centering
    \includegraphics[width=0.48\textwidth]{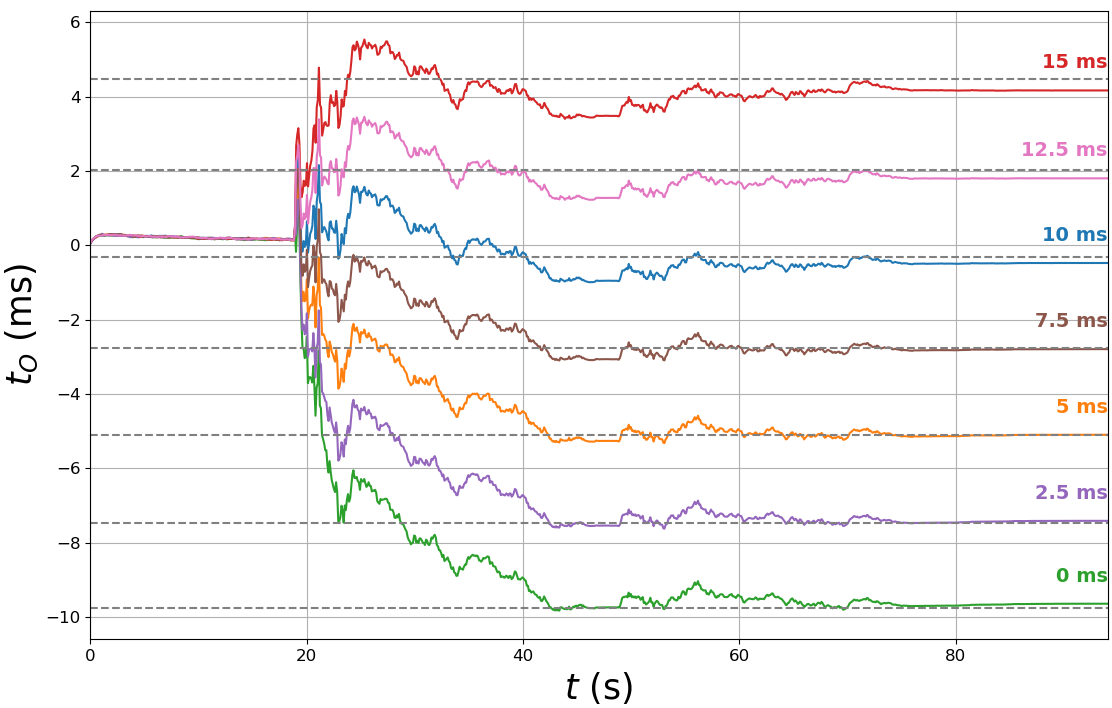}
    \caption{Convergence of the temporal offset in the Gym sequence from the IRS dataset using the proposed method. The radar measurements were artificially shifted forward in time in steps of 2.5\, ms to check whether the proposed method converges correctly to the expected offset values.}
    \label{fig::conv}
    \vspace{-5mm}
\end{figure}

\renewcommand{\tabcolsep}{6pt}
\begin{table*}[t]
\renewcommand{\arraystretch}{1.3}
\noindent\makebox[\textwidth]{
\begin{tabular}{c c c c c c c c c c c c c |c c c c}
 \hline
 \multirow{2}{*}{Method}   & \multicolumn{3}{c}{Gym}  & \multicolumn{3}{c}{Mocap easy}  &  \multicolumn{3}{c}{Mocap medium}  & \multicolumn{3}{c}{Mocap difficult }  & \multicolumn{3}{c}{Mean}   \\  \cline{2-16}
   &   $t_{rel}$ & $r_{rel}$  & ATE & $t_{rel}$ & $r_{rel}$ & ATE & $t_{rel}$ & $r_{rel}$ & ATE & $t_{rel}$ & $r_{rel}$ & ATE & $t_{rel}$ & $r_{rel}$ & ATE \\
 \hline
  EKF-RIO \cite{doer_online} &  0.057 &  0.287 & 1.210 & \textbf{0.030} & 0.351 & 0.193 &  \textbf{0.051} & 0.556 & 0.216 & \textbf{0.033}  & 0.665 &  0.437 & 0.043 & 0.465& 0.514 \\
  
  Graph-RIO \cite{mfi-rik} &  0.032 & 0.428 & 0.318 & 0.035 & 0.564 & 0.153 & 0.068 & 0.849 & 0.148 & 0.045  & 1.114 &  0.461 & 0.045 & 0.739 & 0.270 \\
  \hline
  LC-RIO  &  0.029 &  0.303 & \textbf{0.274} & 0.033 & 0.330 & \textbf{0.124} &  0.066 & 0.481 & \textbf{0.128} & 0.040  & 0.619 &  0.489 & 0.042 & 0.433 & 0.254 \\
  
\textbf{RIO-T}  &  \textbf{0.027} &  \textbf{0.265} & 0.278& \textbf{0.030} & \textbf{0.325} & 0.128 &  0.062 & \textbf{0.464} & 0.130 & 0.035  & \textbf{0.597} &  \textbf{0.420} & \textbf{0.039} & \textbf{0.413} & \textbf{0.239} \\
 \hline
\end{tabular}}
\caption{Relative translational error ($t_{rel}$ [m/m]) and relative rotational error ($r_{rel}$ [deg/m]) per meter, together with the absolute trajectory error [m], calculated with the origin aligned. The best results for each dataset sequence are highlighted in bold, with all values rounded to three decimal places.}
\label{tab::irs}
\vspace{-5mm}
\end{table*}

\subsection{Temporal offset analysis}
To assess the validity of our approach, we tested its convergence on the Gym sequence from the IRS dataset. 
The estimated temporal offset converged to -9.75 ms, suggesting that the IMU experiences a delay with respect to the radar or that the assumption that the trigger timestamp is identical to the timestamps of the radar measurements may not hold.
We suspect that there is a time delay between the sending of the trigger and the actual emission of the electromagnetic waves, as all sensors were hardware synchronized \cite{irs}. 
To validate our estimated temporal offset, we artificially introduced different delays of the radar messages into the dataset (2.5, 5, 7.5, 10, 12.5 and 15 ms) and observed that our method consistently adapts to these offsets, as shown in Fig.~\ref{fig::conv}. 
The gray lines represent the results of our optimization when the entire factor graph was optimized at once. 
The average temporal offset between the online optimized temporal offsets over two consecutive radar-shifted scenarios is 2.37 ms, compared to the ground truth of 2.5 ms, with the largest errors occurring at higher delays. 
This is consistent with the expectation that the assumption that the last acceleration represents the entire state becomes less valid as the time delay increases. 
Nevertheless, our results show that the proposed method provides a robust estimate of the temporal offset.

\begin{figure}[t]
    \centering
    \includegraphics[width=0.48\textwidth]{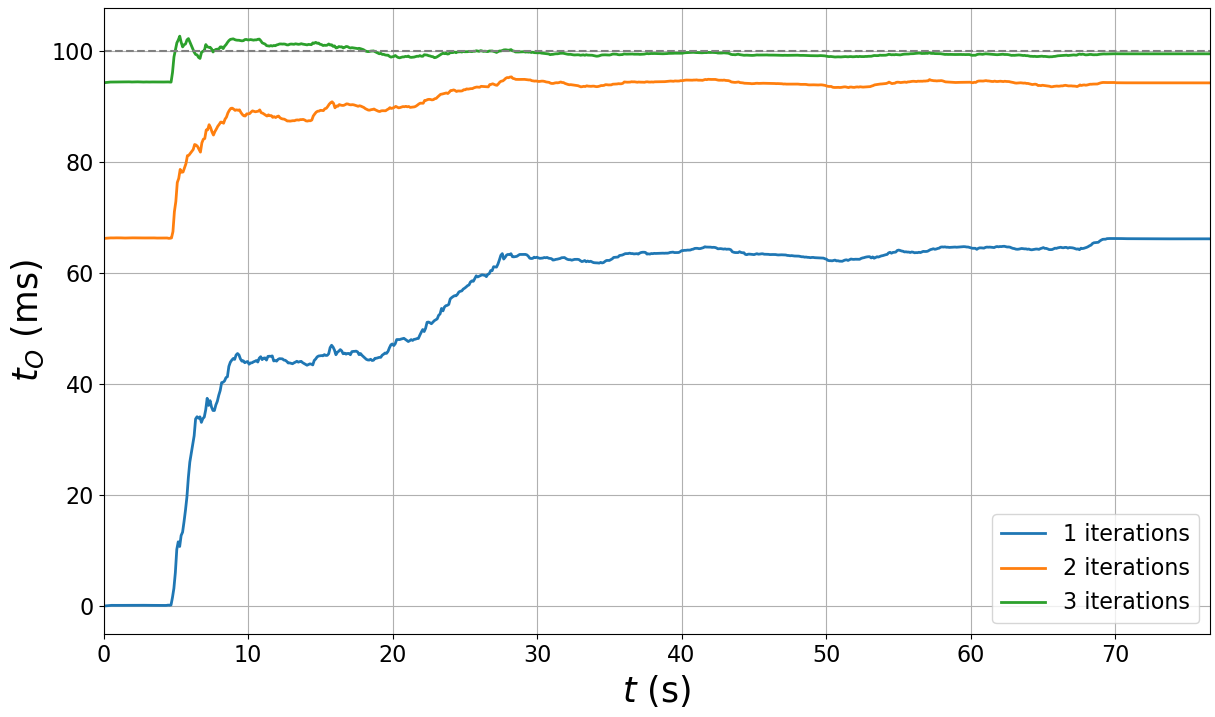}
    \caption{Convergence of the temporal offset in the Mocap difficult sequence from the IRS dataset using the proposed method. The radar measurements were artificially shifted by 112.15 ms to simulate the time delay of a radar of 100 ms. We tested our method over three iterations to evaluate its convergence ability even with large temporal offsets.}
    \label{fig::time3}
    \vspace{-5mm}
\end{figure}
To evaluate the robustness of the proposed method in dealing with significant temporal shifts, we conducted experiments by artificially introducing a significant radar delay of 100 ms. 
Since the estimated delay in the Mocap difficult sequence was 12.15 ms, the total applied delay was set to 112.15 ms, as shown in Figure~\ref{fig::time3}. 
In the first iteration, the estimated temporal delay was 66.2 ms, resulting in an estimation error of 33.8 ms. 
To improve the accuracy of the estimation, we introduced a time shift of 66.2 ms so that the method could use a more recent acceleration measurement. 
This adjustment reduced the estimated temporal offset to 28.1 ms. 
Combined with the previous offset, the cumulative estimate converged to approximately 94.3 ms. 
In the following and final iteration, the method further refined the estimate and achieved an offset of 5.2 ms. 
This resulted in a total estimated delay of approximately 99.5 ms, which exactly matched the specified delay. 
The results show that our method can accurately estimate the delay by iterative refinement even with a significant time offset. 
This approach allows the method to incorporate more recent acceleration measurements over successive iterations, resulting in improved estimation accuracy. 
By using these more recent measurements instead of relying only on the last measurement, the method captures finer temporal dynamics, which further increases the accuracy of the offset estimation. 
Therefore, our method can estimate the temporal offset online when the delay is not significant, and it can approximate larger temporal delays, which can be further refined by iterative estimation to improve the accuracy.

It is important to note that the method used in both experiments to analyze temporal offset did not initially converge for a fixed time period due to the stationary segments at the beginning of the IRS dataset sequences.

\begin{figure}[t]
    \centering
    \includegraphics[width=0.5\textwidth]{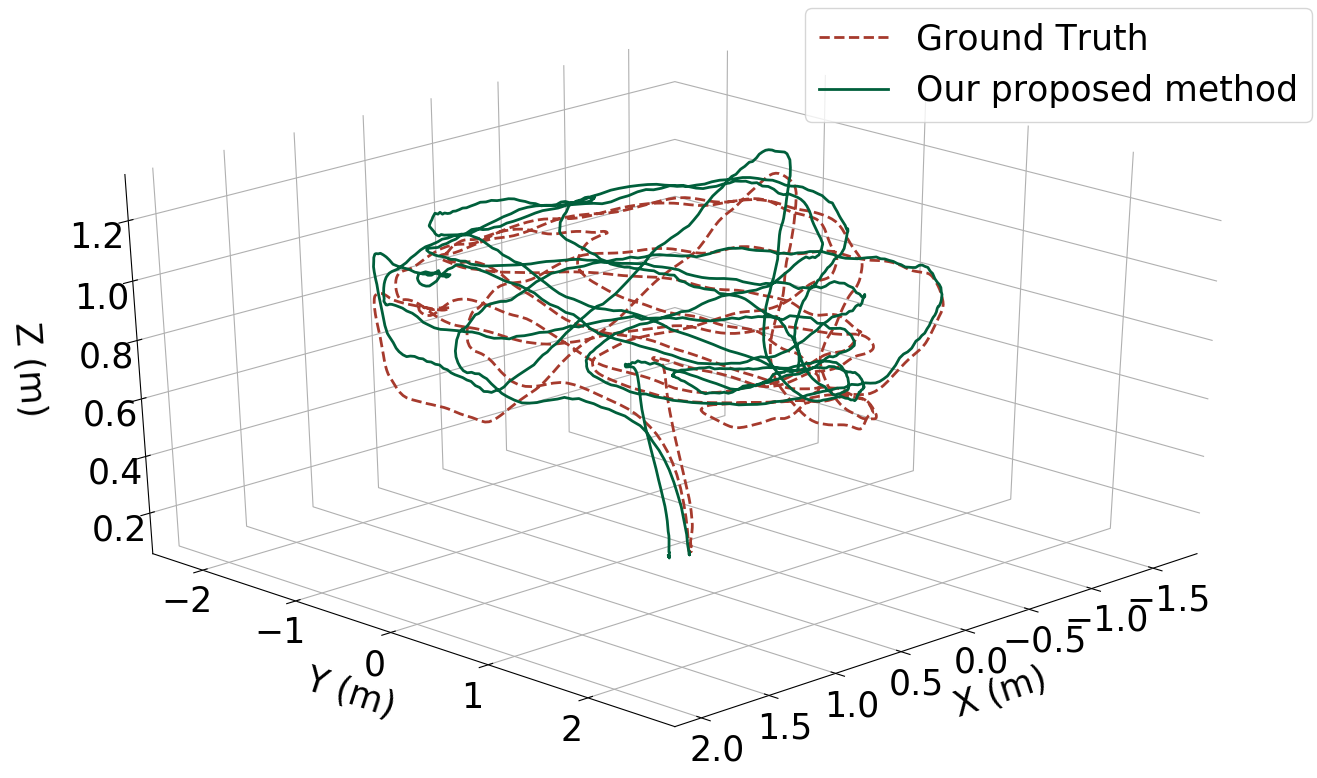}
    \caption{Trajectory visualization of RIO-T on the Mocap easy sequence from the IRS dataset~\cite{irs}.}
    \label{fig::com}
    \vspace{-5mm}
\end{figure}

\renewcommand{\tabcolsep}{8pt}
\begin{table}[t]
\renewcommand{\arraystretch}{1.26}
\centering
\caption{Relative translational error ($t_{rel}$ [m/m]) and relative rotational error ($r_{rel}$ [deg/m]) per meter, together with the absolute trajectory error [m], calculated with an aligned origin. Each radar inertial odometry method was evaluated with both the original and corrected timestamps to simulate the effects of temporal misalignment on odometry accuracy.}
\scriptsize 
\begin{tabular}{c c c c c}
\hline
\textbf{Sequence} & \textbf{Method} & $t_{rel}$ &  $r_{rel}$ & ATE \\ 
\hline
\multirow{2}{*}{Gym} 
    & EKF-RIO \cite{doer_online}          & 0.057  & 0.287  & 1.210  \\
    & EKF-RIO (+9.96 ms)    & \textbf{0.054} & \textbf{0.286}  & \textbf{1.051} \\
    \cline{2-5}
    & Graph-RIO  \cite{mfi-rik}        & 0.032 & 0.428 & 0.318  \\
    & Graph-RIO (+9.96 ms)    & \textbf{0.029}  & \textbf{0.376} & \textbf{0.285}  \\
    \cline{2-5}
    & LC-RIO             & 0.029 & 0.303 & \textbf{0.274} \\
    & LC-RIO (+9.96 ms)       & \textbf{0.028}   & \textbf{0.272} & 0.278 \\ 
\hline
\multirow{2}{*}{Mocap dark} 
    & EKF-RIO \cite{doer_online}          & \textbf{0.092} & \textbf{1.983}   & 0.504  \\
    & EKF-RIO (+11.05 ms)    &  0.094  & 2.042   & \textbf{0.330}   \\ 
    \cline{2-5}
    & Graph-RIO \cite{mfi-rik}          & \textbf{0.092} & 2.031 & 0.582  \\
    & Graph-RIO (+11.05 ms)    & 0.094  & \textbf{1.977}   &  \textbf{0.551} \\
    \cline{2-5}
    & LC-RIO             & \textbf{0.090} & 1.981 & 0.282  \\
    & LC-RIO (+11.05 ms)       & 0.094   & \textbf{1.961}  & \textbf{0.209}   \\ 
\hline
\multirow{2}{*}{Mocap difficult} 
    & EKF-RIO   \cite{doer_online}           & \textbf{0.033} & \textbf{0.665} & 0.437 \\
    & EKF-RIO (+12.15 ms)     & 0.036   & 0.846  & \textbf{0.170}  \\
    \cline{2-5}
    & Graph-RIO  \cite{mfi-rik}         & 0.045 & 1.114 & 0.461  \\
    & Graph-RIO (+12.15 ms)    & \textbf{0.037}  & \textbf{0.964}   & \textbf{0.174}  \\ 
    \cline{2-5}
    & LC-RIO             & 0.040 & 0.619  & 0.489  \\
    & LC-RIO (+12.15 ms)       & \textbf{0.035}   & \textbf{0.550}  &  \textbf{0.267} \\ 
\hline
\hline
\multirow{2}{*}{Mean} 
    & EKF-RIO \cite{doer_online}             & \textbf{0.061} & \textbf{0.978} & 0.717  \\
    & EKF-RIO (-$\Delta t$)    & \textbf{0.061} & 1.058 & \textbf{0.517} \\
    \cline{2-5}
    & Graph-RIO  \cite{mfi-rik}         & 0.056 & 1.191 & 0.454 \\
    & Graph-RIO (-$\Delta t$)    & \textbf{0.053} & \textbf{1.106} & \textbf{0.337}  \\ 
    \cline{2-5}
    & LC-RIO             & 0.053 & 0.968 & 0.348 \\
    & LC-RIO (-$\Delta t$)      & \textbf{0.052} & \textbf{0.928} & \textbf{0.251} \\ 
\hline
\end{tabular}
\label{tab:to}
\vspace{-3mm}
\end{table}

\subsection{Odometry analysis}
We compared our RIO-T approach with three radar-inertial odometry methods: an EKF-based approach (EKF-RIO~\cite{doer_online} without barometer data), a tightly coupled factor graph-based method (Graph-RIO~\cite{mfi-rik}), and a loosely coupled factor graph-based method (LC-RIO). 
The proposed method is denoted as RIO-T.
LC-RIO represents a group of methods \cite{coral,multi_radar_cauchy,riv} that integrate a loosely coupled radar-inertial structure into a factor graph without additional scan registration, point tracking, ground detection, or specific ego-velocity estimation techniques. 
We use the same ego-velocity estimation techniques as RIO-T to reliably investigate the effects of online temporal calibration on the overall accuracy of odometry.

Table ~\ref{tab::irs}  shows the RPE together with the ATE using the origin alignment for four IRS dataset sequences: Gym, Mocap easy, Mocap medium and Mocap difficult. 
LC-RIO and RIO-T were initialized with the same parameters to allow a fair and transparent evaluation of the temporal offset estimation. 
The results from Table ~\ref{tab::irs} show the advantage of integrating the temporal offset into the radar measurements and the state, achieving an accuracy that can compete with other modern radar inertial odometry algorithms and in some cases even surpasses them. 
The effect of estimating the temporal offset is particularly evident in the most dynamic sequence, Mocap difficult, where omitting the offset leads to the largest discrepancies, reducing the ATE by 14.11\%. 
In contrast, online temporal calibration offers only limited benefits in the Mocap easy and Mocap medium sequences, where the movement is less agile. 
The mean RPE and ATE results indicate that the introduction of the temporal offset has a positive effect on the accuracy of odometry, especially in scenarios with fast motion and high acceleration, where the temporal delay is also estimated.

\subsection{Impact of temporal offset on radar-inertial odometry}
Using the estimated temporal offset for each sequence, we artificially delayed the radar time stamps and evaluated the state-of-the-art ego-motion estimators both with and without delay to analyze the effects of temporal delay on the accuracy of odometry. 
The results in Table~\ref{tab:to} clearly show the significant influence of the correction of the temporal offset. 
In particular, EKF-RIO, Graph-RIO, and LC-RIO achieved 27.89\%, 25.77\%, and 27.87\% lower ATE, respectively, when using the corrected timestamps compared to the original timestamps. 
These results emphasize the critical role of temporal calibration in radar IMU systems to improve the accuracy of odometry in dynamic scenarios when there is a time offset between the measurements of the two sensors.

\subsection{Discussion}
The main focus of this work is to demonstrate the impact of explicit modeling of temporal offset on the accuracy of radar-based odometry and its general importance for downstream tasks such as odometry. 
Our results show that integrating the temporal offset into the estimation process significantly improves the performance of odometry compared to approaches that assume perfect synchronization. 
This improvement is particularly evident in sequences characterized by high acceleration and rapid motion changes, where the temporal offset between radar and IMU measurements leads to cumulative errors over time. 
This effect is most evident in the difficult Mocap sequence, where ignoring the temporal offset resulted in the largest discrepancy. 
In contrast, our method successfully compensates for this offset, resulting in a more accurate trajectory estimate, which is especially crucial for agile movements. 
Although the IRS dataset uses hardware synchronization, we emphasize that the impact of temporal offset is likely to be even greater in scenarios where such synchronization is unavailable or impractical. 
Therefore, we have shown that our method can iteratively converge even in the presence of significant time delays by integrating more recent radar measurements, thus improving the accuracy of the estimated offset. 
In practical applications where radar timestamps are not hardware-triggered, our method can serve as an important tool to improve accuracy without requiring expert knowledge of low-level hardware synchronization techniques.

Given the positive effect on odometry accuracy obtained by assuming constant acceleration around the radar ego-velocity measurement in a loosely coupled manner, the same principle can be applied in a tightly coupled implementation, where the measurement model operates directly on Doppler measurements instead of estimated ego-velocities.
While our current approach assumes a constant acceleration model to estimate the offset, future work will explore more advanced models that incorporate a wider range of previous IMU measurements onto the radar measurement factor, rather than relying solely on the last measurement value.
This would enable a more accurate representation of acceleration, leading to a more refined and theoretically grounded temporal calibration.

Furthermore, following the approaches proposed in~\cite{jan_online, doer_online}, we plan to incorporate an extrinsic online calibration to test whether both spatial and temporal calibration can converge simultaneously and to evaluate their combined effect on the accuracy of odometry.

\vspace{-0.5mm}
\section{Conclusion}
\label{sec:conclusion}
\vspace{-0.5mm}
In this paper, we have presented a novel radar-inertial odometry system based on the factor graph optimization with online calibration of the temporal offset for radar measurements. 
The results show that the estimation and compensation of the temporal offset significantly improves the accuracy of the odometry. 
The convergence of the estimated offset was validated using artificially simulated delays on real-world data and the proposed method was evaluated in comparison to state-of-the-art algorithms. 
We also compared the performance of state-of-the-art radar inertial ego-motion estimators with and without temporal offset and confirmed the positive influence of the estimated time delay on the accuracy of odometry. 
Furthermore, we have shown that even with hardware triggering of the radar, the sensors may still exhibit a delay or the assumption that the radar data are perfectly synchronized with the trigger may not always hold. 
Our results indicate that temporal misalignment is particularly important for agile movements and can severely affect the performance of odometry.

Future work will focus on refining the temporal offset estimation using more complex representations and extending our approach to include extrinsic online calibration.

\addtolength{\textheight}{-5cm}   




\section*{ACKNOWLEDGMENT}
This research has been supported by the European Regional Development Fund under the grant PK.1.1.02.0008 (DATACROSS).


\vspace{-1mm}
\balance
\bibliography{thebibliography}
\bibliographystyle{IEEEtran}

\end{document}